\documentclass{isprs}
\usepackage{subfigure}
\usepackage{setspace}
\usepackage{geometry} 
\usepackage{epstopdf}
\usepackage{amsmath,epsfig,mathrsfs,amssymb}
\usepackage[labelsep=period]{caption}  
\usepackage[british]{babel} 
\usepackage{booktabs}
\usepackage{multirow}
\usepackage{graphicx}

\newcommand{\norm}[1]{\lVert#1\rVert}

\DeclareMathOperator{\T}{T}

\begin{document}

\title{Spatial-Spectral Manifold Embedding of Hyperspectral Data}

\author{
  D. Hong\textsuperscript{1,2}, J. Yao\textsuperscript{1,3,4}, X. Wu\textsuperscript{5}, J. Chanussot\textsuperscript{6}, X. Zhu\textsuperscript{1,3}}

\address{
	\textsuperscript{1} Remote Sensing Technology Institute (IMF), German Aerospace Center (DLR), 82234 Wessling, Germany\\
	- (danfeng.hong, jing.yao, xiaoxiang.zhu@dlr.de)\\
	\textsuperscript{2}  Univ. Grenoble Alpes, CNRS, Grenoble INP, GIPSA-lab, 38000 Grenoble, France\\
	\textsuperscript{3} Signal Processing in Earth Observation (SiPEO), Technical University of Munich (TUM), 80333 Munich, Germany\\
	\textsuperscript{4} School of Mathematics and Statistics, Xi'an Jiaotong University, 710049 Xi'an, China\\
	- jasonyao@mail.xjtu.edu.cn\\
	\textsuperscript{5} School of Information and Electronics, Beijing Institute of Technology (BIT), 100081 Beijing, China.\\
	- 040251522wuxin@163.com\\
	\textsuperscript{6} Univ. Grenoble Alpes, INRIA, CNRS, Grenoble INP, GIPSA-lab, 38000 Grenoble, France\\
	- jocelyn@hi.is\\
}

\icwg{}   

\abstract{
\textit{This is a pre-print version accepted for publication in the International Archives of the Photogrammetry, Remote Sensing and Spatial Information Sciences.}

In recent years, hyperspectral imaging, also known as imaging spectroscopy, has been paid an increasing interest in geoscience and remote sensing community. Hyperspectral imagery is characterized by very rich spectral information, which enables us to recognize the materials of interest lying on the surface of the Earth more easier. We have to admit, however, that high spectral dimension inevitably brings some drawbacks, such as expensive data storage and transmission, information redundancy, etc. Therefore, to reduce the spectral dimensionality effectively and learn more discriminative spectral low-dimensional embedding, in this paper we propose  a novel hyperspectral embedding approach by simultaneously considering spatial and spectral information, called spatial-spectral manifold embedding (SSME). Beyond the pixel-wise spectral embedding approaches, SSME models the spatial and spectral information jointly in a patch-based fashion. SSME not only learns the spectral embedding by using the adjacency matrix obtained by similarity measurement between spectral signatures, but also models the spatial neighbours of a target pixel in hyperspectral scene by sharing the same weights (or edges) in the process of learning embedding. Classification is explored as a potential strategy to quantitatively evaluate the performance of learned embedding representations. Classification is explored as a potential application for quantitatively evaluating the performance of these hyperspectral embedding algorithms. Extensive experiments conducted on the widely-used hyperspectral datasets demonstrate the superiority and effectiveness of the proposed SSME as compared to several state-of-the-art embedding methods.
}


\maketitle

\section{Introduction}
Currently operational hyperspectral missions, such as DLR Earth Sensing Imaging Spectrometer (DESIS) \cite{krutz2018desis}, Gaofen-5 \cite{ren2017improving}, Environmental Mapping and Analysis Program (EnMAP) \cite{guanter2009simulation}, enable the recognition and identification of the materials of interest at a more accurate level compared to the multispectral data \cite{hong2015novel} or RGB data \cite{wu2018msri,wu2019orsim}. However, due to the effects of \textit{curse of dimensionality}, some drawbacks are inevitably introduced with the high spectral dimensionality, possibly leading to the degradation of spectral information. As a result, the dimensionality reduction is a necessary step before the high-level data analysis is performed.

\begin{figure*}[!t]
	  \centering
		\subfigure{
			\includegraphics[width=1\textwidth]{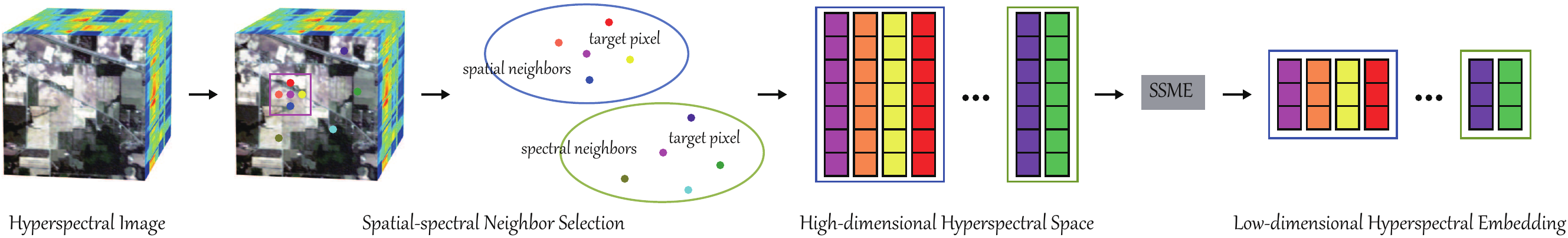}
		}
        \caption{An illustration for the holistic workflow of the proposed SSME model.}
\label{fig:workflow}
\end{figure*}

Over the past decades, a large amount of dimensionality reduction approaches have been successfully applied in many computer vision related fields, such as low-level vision analysis \cite{bi2017unsupervised,kang2020learning}, biometric \cite{hong2014improved,hong2016robust}, large-scale data classification \cite{hong2016k,huang2019multi,bi2018graph,huang2020deep,bi2019active}, multimodal data analysis \cite{zhang2019estimation,zhang2019land,hong2020learning}, data fusion \cite{hu2019comparative,hu2019mima}, etc. Among them, spectral manifold embedding, as a popular topic in hyperspectral dimensionality reduction \cite{hong2016local}, has attracted a growing attention in various hyperspectral remote sensing applications, such as hyperspectral image denoising \cite{cao2018tensor,cao2018robust}, land cover and land use classification \cite{hang2019cascaded,hong2019learnable}, spectral unmixing \cite{hong2018sulora,hong2019augmented,yao2019nonconvex}, target detection and recognition \cite{li2018real,wu2019fourier}, and multimodal data analysis \cite{liu2019stfnet,hong2019cospace,hang2020classification}. It is well known that the hyperspectral imagery is a three-dimensional imaging product by continuously scanning the region of interest (ROI) to obtain hundreds or thousands of two-dimensional images finely sampled from the wavelength nearly covering the whole electromagnetic spectrum, e.g., 300nm to 2500nm. This enables the identification and detection of materials lying on the surface of the Earth at a more accurate level compared to other optical data, e.g., RGB. In the meanwhile, high dimensional spectral signatures are also introduced some serious drawbacks. For example, high storage and computational cost, redundant information, and complex noises caused by atmospheric correlation would have a negative influence on the spectral discrimination of hyperspectral images, further degrading the performance of high-level applications, e.g., classification, detection.

Recently, enormous effects have been made to enhance the quality of low dimensional hyperspectral embedding in the spectral domain. More specifically, Hong \textit{et al.} \cite{hong2016local} proposed robustly select the neighbouring pixels for hyperspectral dimensionality reduction. The same investigators in \cite{hong2017learning} further design a novel hyperspectral low-dimensional embedding algorithm to learn a robust local manifold representation for dimensionality reduction of hyperspectral images. Inspired by the recent success of deep learning \cite{wu2020frft}, Hong \textit{et al.} \cite{hong2018joint} developed a joint and progressive learning strategy to learn the low-dimensional representations by using manifold regularization techniques in each layer. The proposed deep embedding model has demonstrated its superiority and effectiveness in the hyperspectral dimensionality reduction task.

Yet the spatial information \cite{hong2020invariant} is less investigated by the researchers who are working in the remote sensing community in the process of hyperspectral embedding \cite{rasti2020feature}. It is well known that the spatial information has been proven to be effective in the hyperspectral image classification task \cite{cao2020an}, owing to the important and reasonable assumption in hyperspectral images, that is, the target pixel and its neighboring pixels would share the same category to a great extent. We have to admit, however, that the spatial information modeling is capable of improving the discriminative ability of learned embedding representations more effectively, as the spatial structure is one of most important physically meaningful properties in hyperspectral imaging.

For the aforementioned reason, in this paper we attempt to develop a novel hyperspectral embedding approach by simultaneously considering spatial and spectral information, called spatial-spectral manifold embedding (SSME). SSME not only learns the spectral embedding by using the adjacency matrix obtained by similarity measurement between spectral signatures, but also models the spatial neighbours of a target pixel in hyperspectral scene by sharing the same weights (or edges) in the process of learning embedding. Classification is explored as a potential strategy to quantitatively evaluate the performance of learned embedding representations. Extensive experiments conducted on the widely-used hyperspectral datasets demonstrate the superiority and effectiveness of the proposed SSME as compared to several state-of-the-art embedding methods in terms of overall accuracy (OA), average accuracy (AA), and kappa coefficient ($\kappa$). More specifically, our contributions of this paper can be highlighted as follows:
\begin{itemize}
    \item A novel hyperspectral dimensionality reduction approach -- spatial-spectral manifold embedding (SSME) -- is devised to learn the low-dimensional manifold embedding of the hyperspectral data.
    \item Beyond the pixel-wise spectral embedding, we propose to construct the spatial-spectral weight matrix in spectral embedding, yielding more smooth low dimensional hyperspectral embedding.
    \item Experimental results conducted on a widely-used hyperspectral data demonstrate the effectiveness and superiority of the proposed SSME approach.
\end{itemize}

The rest of this paper is organized as follows. Section 2 details the methodology of the proposed SSME approach with some necessary formulation derivation. Accordingly, extensive experiments are conducted in comparison with several competitive methods in Section 3. Finally, we draw a conclusion in Section 4.

\section{Methodology}\label{Methodology}

Manifold embedding, also known as manifold learning, is built on the graph embedding framework \cite{yan2006graph} by attempts to capture the underlying structure of the original data and preserve it in the latent embedding space. The embedding process mainly consists of three steps in the following.
\begin{itemize}
    \item Neighbor selection on the spectral domain by spectrally measuring the similarities between pixels;
    \item Adjacency matrix computation between each target pixel and its neighbouring pixels by using regression-based methods \cite{hong2019regression} or Gaussian kernel functions;
    \item Calculation of embedding by solving a generalized eigen-decomposition problem. 
\end{itemize}

Unlike the previous manifold embedding techniques, such as locally linear embedding (LLE) \cite{roweis2000nonlinear}, Laplacian eigenmaps (LE) \cite{belkin2002laplacian}, and their linearized approaches: locality persevering projections (LPP) \cite{he2004locality} and neighborhood preserving embedding (NPE) \cite{he2005neighborhood}, that only conduct on the spectral domain, the newly-proposed spatial-spectral manifold embedding (SSME) performs the low-dimensional embedding process from both spatial and spectral domains in a joint fashion. Similarly, SSME follows the graph embedding framework as well. Given a hyperspectral image $\mathbf{X}\in \mathcal{R}^{D\times N}$ with $D$ bands by $N$ pixels, $\mathbf{x}_{i,j}$ is denoted as the spectral signature located in $(i,j)$ of the image. We then have
\begin{itemize}
    \item Spatial-spectral neighbor selection by using spatially prior knowledge and Euclidean distance based similarity measurement in spectral domains, respectively, which can be written as follows
    \begin{equation}
    \label{eq1}
    \begin{aligned}
           \phi_{i}^{spa}\leftarrow[\mathbf{x}_{i,j},\mathbf{x}_{i-1,j},\mathbf{x}_{i,j-1},\mathbf{x}_{i+1,j},\mathbf{x}_{i,j+1}],\\
           \phi_{i}^{spe}\leftarrow sort\{\{\norm{\mathbf{x}_{i}-\mathbf{x}_{k}}_{2}\}_{i=1}^{N}\}_{k=1}^{N}.
    \end{aligned}
    \end{equation}
    $\phi_{i}^{spa}$ and $\phi_{i}^{spe}$ denote the spatial and spectral neighbours of the target pixel $\mathbf{x}_{i}$, where the latter one can be obtained by sorting the Euclidean distances ($sort$).
    \item Spatially-induced adjacency matrix generation by computing the regression coefficients or weights between the target pixels and their spatial-spectral neighbours. The process can be formulated by
    \begin{equation}
    \label{eq2}
    \begin{aligned}
           &\mathop{\min}_{\mathbf{w}_{i,0}}\sum_{j\in \phi_{i}^{spa}}\norm{\mathbf{x}_{i,j}-\sum_{k\in \phi_{i}^{spe}}\mathbf{x}_{i,k}w_{i,k,j}}_{2}^{2} \\
           &{\rm s.t.} \; \norm{\sum_{k\in \phi_{i}^{spe}}\mathbf{x}_{i,k}(4w_{i,k,0}-\sum_{k=1}^{4}w_{i,k,j})}_{2}^{2}\leq \eta, \\
           &\qquad \mathbf{w}_{i,j}^{\T}\mathbf{w}_{i,j}=1,
    \end{aligned}
    \end{equation}
    where $\mathbf{x}_{i,k}\in \phi_{i,k}^{spe}$ represents the $k$ nearest neighbors selected from the spectral domain, and $j\in \phi_{i}^{spa}$ is defined as the target pixel in the HSI and its neighbouring pixels, respectively. Accordingly, $\mathbf{w}_{i,j}=[w_{i,1,j},...,w_{i,k,j},...], j\in \phi_{i}^{spa}$, where $\mathbf{w}_{i,0}$ is the to-be-estimated regression coefficients of the target pixel, and $\eta$ is the tolerate error ($10^{-3}$ in our case). 
    
    With the estimated $\mathbf{w}$, the affinity weights $\mathbf{A}$ can be obtained by using the following equation:
    \begin{equation}
    \label{eq3}
    \begin{aligned}
          \mathbf{A}_{i,0,k}=
            \begin{cases}
            \mathbf{w}_{i,0,k}+\mathbf{w}_{k,i,0}-\mathbf{w}_{i,0,k}\mathbf{w}_{k,i,0}, \;\; k\in \phi_{i}^{spe},\\
             0, \;\; {\rm otherwise}.
            \end{cases}
    \end{aligned}
    \end{equation}
    \item Joint embedding guided by the aforementioned adjacency matrix by solving a generalized eigen-decomposition problem. Once the affinity matrix $\mathbf{A}$ is given, the final hyperspectral embedding $\mathbf{Y}=\{\mathbf{y}_{i=1}^{N}\}$ can be computed by solving the minimization problem as follows.
    \begin{equation}
    \label{eq4}
    \begin{aligned}
            &\mathop{\min}_{\mathbf{Y}}\sum_{i=1}^{N}\norm{\mathbf{y}_{i}-\sum_{k\in \phi_{i,k}^{spe}}\mathbf{A}_{i,k}\mathbf{y}_{k}}_{2}^{2}, \\
            & {\rm s.t.} \;\sum_{i=1}^{N}\mathbf{y}_{i}=0, \;\; \frac{1}{N}\sum_{i=1}^{N}\mathbf{y}_{i}\mathbf{y}_{i}^{T}=\mathbf{I}.
    \end{aligned}
    \end{equation}
\end{itemize}

Fig. \ref{fig:workflow} illustrates a workflow of our proposed SSMR method for extracting the low-dimensional hyperspectral embedding representations.

\section{Experiments}
\subsection{Data Description}

To assess the effectiveness of the proposed SSME in hyperspectral embedding task, classification is selected to be a potential strategy \cite{Gao2020Spectral}. In our case, a simple but efficient classifier: nearest neighbour (NN), is used. Moreover, as a widely-used and classic hyperspectral dataset, the Indian pines scene is chosen in our experiments. It was acquired by the Airborne Visible Infrared Imaging Spectrometer (AVIRIS) sensor in the northwestern of Indiana, USA, which consists of $145\times 145$ pixels with 220 spectral bands covering the wavelength from 400$nm$ to 2500$nm$. There are 16 classes in the studied scene.  A fixed and popular training and test sets widely used in many references \cite{hong2019learning} is given in Table \ref{Table:TRTE}, and a false-color image of the hyperspectral data is shown in the first place of Fig. \ref{fig:CM_IndinePines}.

\begin{table}[!t]
\centering
\caption{Class names as well as the number of training and test samples for each class on the hyperspectral dataset over the area of Indine Pines.}
\resizebox{0.45\textwidth}{!}{
\begin{tabular}{c||c|c|c}
\toprule[1.5pt]
No. & Class Name & Training Samples & Test Samples\\
\hline \hline Class 1&CornNotill&50&1384\\
Class 2&CornMintill&50&784\\
 Class 3&Corn&50&184\\
 Class 4&GrassPasture&50&447\\
 Class 5&GrassTrees&50&697\\
 Class 6&HayWindrowed&50&439\\
 Class 7&SoybeanNotill&50&918\\
 Class 8&SoybeanMintill&50&2418\\
 Class 9&SoybeanClean&50&564\\
 Class 10&Wheat&50&162\\
 Class 11&Woods&50&1244\\
 Class 12&BuildingsGrassTrees&50&330\\
 Class 13&StoneSteelTowers&50&45\\
 Class 14&Alfalfa&15&39\\
 Class 15&GrassPastureMowed&15&11\\
 Class 16&Oats&15&5\\
\hline \hline &Total&695&9671\\
\bottomrule[1.5pt]
\end{tabular}
}
\label{Table:TRTE}
\end{table}

\subsection{Results}\label{Results and Analysis}

\begin{table}[!t]
\centering
\setlength{\abovecaptionskip}{5pt}
\setlength{\belowcaptionskip}{5pt}
\caption{Classification performance comparison between different algorithms on the Indian Pines dataset. The best results are shown in bold. Furthermore, the values behind the name of algorithms mean the dimensions used in these compared methods.}
\resizebox{0.48\textwidth}{!}{
\begin{tabular}{c|cccc|c}
\toprule[1.5pt]
Methods & OSF (220) & PCA (30) & LE (60) & LLE (60) & SSME (16)\\
\hline \hline
       OA & 65.89 & 65.76 & 64.45 & 69.05& \bf 76.46\\
       AA & 75.71 & 75.70 & 74.84 & 76.94 & \bf 86.67\\   
       $\kappa$ & 0.6148 & 0.6138 &0.5997 & 0.6500 & \bf 0.7350\\
\hline \hline
Class 1 & 51.66	& 51.81 & 50.51 & 56.00 & \bf 71.97\\
Class 2 & 57.40 & 57.91 & 56.51 & 57.14 & \bf 78.95\\
Class 3 & 70.65	& 69.57 & 67.39 & 70.11 & \bf 90.22\\
Class 4 & 88.14	& 88.37	& 87.92	& 90.83 & \bf 92.39\\
Class 5 & 81.78	& 81.78 & 80.92 & \bf 90.82 & 71.45\\
Class 6 & 95.90 & 96.13 & 94.53 & 95.44 & \bf 98.63\\
Class 7 & 66.56	& 67.97	& 68.30	& 68.52	& \bf 84.31\\
Class 8 & 55.21	& 54.22	& 52.11	& 59.10	& \bf 59.26\\
Class 9 & 53.01	& 52.48	& 50.53	& 59.40 & \bf 84.75\\
Class 10 & 98.15 & 98.15 & 97.53 & \bf 99.38 & 98.15\\
Class 11 & \bf 82.88 & 82.48 & 80.95 & 81.11 & 82.07\\
Class 12 & 50.91 & 51.21 & 51.21 & 66.06 & \bf 93.64\\
Class 13 & \bf 97.78 & \bf 97.78 & \bf 97.78 & 93.33 & \bf 97.78\\
Class 14 & 79.49 & 79.49 & 79.49 & 82.05 & \bf 92.31\\
Class 15 & 81.82 & 81.82 & 81.82 & 81.82 & \bf 90.91\\
Class 16 & \bf 100.00 &	\bf 100.00 & \bf 100.00 & 80.00 & \bf 100.00\\
\bottomrule[1.5pt]
\end{tabular}
}
\label{tab:IndianPine}
\end{table}

\begin{figure*}[!t]
	  \centering
		\subfigure{
			\includegraphics[width=1\textwidth]{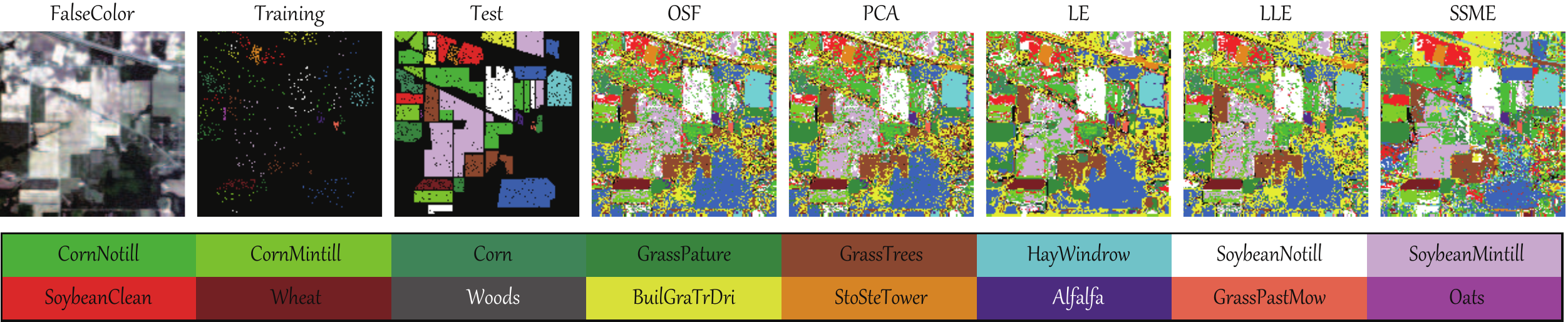}
		}
        \caption{Classification maps obtained by different hyperspectral embedding approaches on the Indine Pines dataset.}
\label{fig:CM_IndinePines}
\end{figure*}

A false-color image of this scene is shown in the first column of Fig. \ref{fig:CM_IndinePines} and the distribution of training and test sets are also given in the subsequent column of Fig. \ref{fig:CM_IndinePines}. 

Several hyperspectral embedding baselines are selected to evaluate the quality of learned embedding representations using the different methods, such as the original spectral features (OSF), principal component analysis (PCA) \cite{wold1987principal}, LE, LLE, and ours (SSME). Quantitative classification accuracies of these compared methods in terms of OA, AA, and $\kappa$ are listed in Table \ref{tab:IndianPine}, while Fig. \ref{fig:CM_IndinePines} visualizes the classification maps with false-color image and the distributions of training and test sets.

Overall, the results using PCA are basically consistent with those using OSF in all indices. Also, LE holds similar embedding results assessed by means of classification tasks compared to PCA and OSF. By linearly regressing the local neighboring relationship of a target pixel, LLE performs better than the aforementioned embedding methods at an increase of about 5\% OAs. As expected, our proposed SSME outperforms others dramatically by jointly considering spatial and spectral information, showing the superiority in hyperspectral low-dimensional embedding tasks. In addition, the accuracies for most of class using our proposed SSME are higher than those using other competitors, as listed in Table \ref{tab:IndianPine}.

\section{Conclusion}
In this paper, we propose a novel spatial-spectral hyperspectral embedding approach, called spatial-spectral manifold embedding (SSME), for hyperspectral dimensionality reduction in remote sensing community. SSME not only utilizes the spectral information but also modals the spatial information when calculating the low-dimensional embedding. We have to admit, however, that although the SSME is capable of extracting the hyperspectral features well, yet the discriminative ability for feature representations still remains limited due to the relatively weak data fitting ability (linearized). To this end, we would like to introduce more powerful techniques, e.g., deep learning, to further enhance the representation ability in extracted low-dimensional embedding or introduce additional data sources to better guide the hyperspectral embedding, e.g., light detection and ranging (LiDAR) \cite{huang2019semantic}, synthetic aperture radar (SAR) \cite{hu2019topological}, multispectral data, in the future work.
{
	\begin{spacing}{1.17}
		\normalsize
		\bibliography{ISPRSguidelines_authors} 
	\end{spacing}
}


\end{document}